# Human Shape Variation - An Efficient Implementation using Skeleton

Dhriti Sengupta[1], Merina Kundu[2], Jayati Ghosh Dastidar[3]

## Abstract

*It is at times important to detect human presence automatically in secure environments. This needs a shape recognition algorithm that is robust, fast and has low error rates. The algorithm needs to process camera images quickly to detect any human in the range of vision, and generate alerts, especially if the object under scrutiny is moving in certain directions. We present here a simple, efficient and fast algorithm using skeletons of the images, and simple features like posture and length of the object.*

## Keywords

## 1. Introduction

Detection of an object and its movement is a challenging problem in computer image processing. The different shapes of different objects, the different shapes of same the object, and also the changing background environment of the image induce variations in the detection of an object class. Recent approaches in the literature can be roughly classified into: (a) image appearance-based [1, 2], (b) shape driven [3, 4, 5, 6] and (c) mixture of shape and appearance based [7, 8, 9] models. Our work is focused on shape-posture-position based representation of the objects and its corresponding environment. We use the fact that a static camera takes pictures from a fixed angle, and therefore, always captures images which has a static background. Subtracting this background from successive images generates what we can call the foreground, which will consist of dynamic objects coming into view. We then process these dynamic objects. First, we extract its skeleton [10], which is an important shape descriptor.

However, skeletons are sensitive to the boundary deformation and therefore, recognizing objects from skeletons is known to be a difficult task [11].

In fact, the skeletons for two similar objects can vary widely [12]. To tackle this issue, we use two simple heuristic feature elements of a human skeleton. First, we compute its height to width ratio. This gives an estimate of the subject's overall shape. Secondly, we determine the major points on the skeleton where branches begin, thus giving us an idea about the object's important local shape features. Both of these are extremely easy to calculate, and can be done very fast. We use these two features to generate a score for the object, indicating whether the object is likely to be a human or not. Finally, we look at the object's relative positions across frames and decide if it is moving. The paper is organized as follows. Section 2 describes some related works in this field. Section 3 gives the algorithm for static and dynamic image processing. Section 4 describes processing of the skeletons. In Section 5 some experimental data have been presented. Finally, Section 6 concludes.

## 2. Literature Review

Authors in [10, 11, 12, 13, 14, 15, 16, 17, 18] have thoroughly studied the use of skeleton in modelling the object shape on the basis of matching different shapes, which have already been extracted from the images. Some shape matching algorithms such as Shape Contexts [3], Chamfer matching [19], Inner Distance [20], and Data-driven EM [21] are important works in this field. Shape-based approaches have the advantage of being relatively robust against illumination and appearance change. Authors in [9, 22, 23] have worked in the field of matching-based object detection algorithms to decompose a given shape into a group of different parts, and match the resulting parts to edge segments in a given edge image. In the Active Basis algorithm, proposed by the authors in [2] the primary focus is on learning effective appearance models for the object parts by slightly perturbing their locations and orientations.

Our algorithm is based on skeletons of different objects and their differences. This algorithm is

Manuscript received March 07, 2014.
**Dhriti Sengupta**, St. Xavier's College, Kolkata (Autonomous), India.
**Merina Kundu,** St. Xavier's College, Kolkata (Autonomous), India.
**Jayati Ghosh Dastidar**, St. Xavier's College, Kolkata (Autonomous), India.

efficient because it models rigid, as well as non-rigid objects. In this paper, we have represented the configuration of the skeleton using some feature points. Object parts are then identified from these features in the skeleton. In the detection stage, we have determined if the overall posture and shape of the skeleton matches to a human or not, depending on the measurements calculated from the feature points. Thus, our algorithm is simpler and computationally efficient. We have illustrated this algorithm with an example which has given us promising results.

## 3. Processing of images

**A. Collecting information**
We collected static background information of a particular region at different times of the day with different illumination conditions (for example: day, evening and night) and without any other dynamic entity. Stream of real images, which we will call '*frames*', were also collected, and later used for the detection of any dynamic object over the background. We used a frame rate of 10 frames per second for good observation and also to reduce time complexity. In the next stage the following problems were addressed:
  a) Identify what else had appeared over the static background information that was collected.
  b) Determine whether the change had occurred in the background or the foreground.

**B. Detecting Change and Extracting Object**
We compared each of the frames to the static images to find out if any change has occurred. To detect any change between real time frames and the static background image we calculated the correlation coefficient between each of these frames to the static background image using the following formula

$$r = \frac{\sum_m \sum_n (A_{mn} - \bar{A})(B_{mn} - \bar{B})}{\sqrt{\left(\sum_m \sum_n (A_{mn} - \bar{A})^2\right)\left(\sum_m \sum_n (B_{mn} - \bar{B})^2\right)}}$$

where, *r* denotes the correlation coefficient between A and B, where A and B are image matrices of the same size, m×n, $A_{mn}$ is the element of A at position (*m, n*). $\bar{A}$ is the mean value in A, and $\bar{B}$ is the mean value in B. If correlation coefficient of two images is equal to 1 then there is no difference between those two images. We used a threshold value of 0.95 for *r*.

If *r* was less than 0.95, we compared the images pixel-wise, and created a new image called *DIFF*. Every non-zero difference was set to white, and every zero value was maintained. Thus *DIFF* gave a black and white image with only the difference highlighted in white.

We explain this with an example. Consider the following images in Figure 1; the first one being the background, and the second one being a frame shot. The correlation was less than 0.95, so a pixel wise difference was taken and *DIFF* was found, shown in Figure 2.

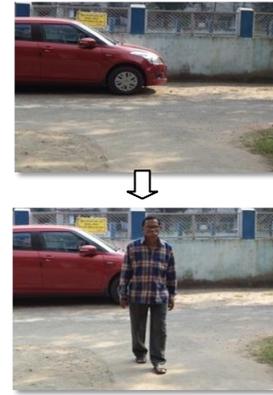

**Figure 1: Two frames, one the static background and the second with a human presence.**

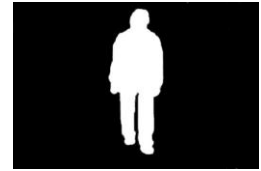

**Figure 2: Dynamic object extraction by differencing**

## 4. Processing of skeletons

**A. Extracting features**
To recognize the object, features of the object are needed. These features preferably should be easy to compute, and easy to manipulate. We have used the *skeleton*, which is the set of points that are equidistant from the nearest edges of the image [10]. There are two main advantages of using skeletons for detection of the object class – (a) it emphasizes the geometric and topological properties of the shape and (b) it retains connectivity in the image (figure 3).

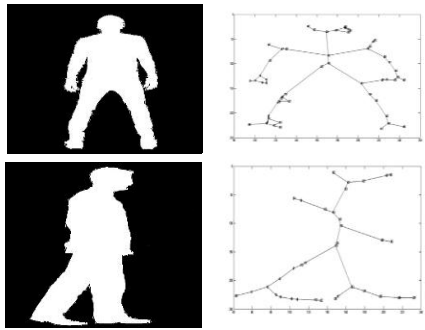

**Figure 3. Examples of skeletons of 2D images.**

**B. Features from Skeletons**
We call a skeleton point having only one adjacent point an **endpoint**; a skeleton point having more points a skeleton **branch**. Every point which is not an endpoint than two adjacent points a **fork point**; and a skeleton segment between two skeletons or a fork point is called a **branch point** (figure 4).

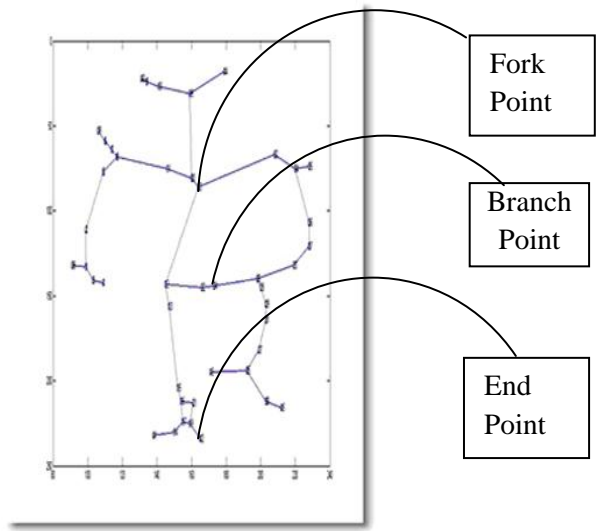

**Figure 4. Skeleton with endpoint, branch point and fork points.**

Skeletons of two dimensional objects often show a lot of spurious edges and branches because of image noise. We used the skeleton generation algorithm using discrete curve evolution [13], to prune such branches. Once we got the clean skeleton, we processed it to find the shape information. Note that the shape of a human being is primarily determined by its relative positions of limbs (arms and legs), neck and head. Thus, if we look at the points where the skeleton is "broken" into forks – we can get a fair idea of where the neck, head, arms, or legs are. This will be very different from other living beings. Here, we have given an example of a processed picture of a dog and the corresponding skeleton (figure 5). The shape of the skeleton and the position of fork points are quite different from that of a human.

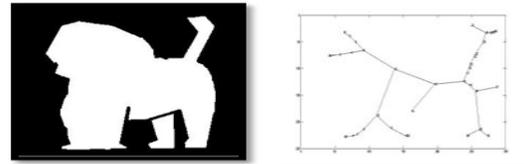

**Figure 5. Skeleton of a dog.**

**C. Detection**
The choice of the inference algorithm critically decides the quality of an object detector. For generic object detection, deeper algorithms are needed, but they also suffer from other drawbacks. Object detection using Markov chain, Monte-Carlo is often slow, though it guarantees to find the global optimal asymptotically. Local pruning procedures are used in a bottom-up and top down process, resulting in quite complex algorithms. Our objective is to find a quick and simple algorithm, given the fact that we have a very specific requirement. We focused on the shape-posture position of the object and then decided the direction of the movement depending on the relative positions of the dynamic object in adjacent frames. Our algorithm relies on certain features of skeletons of different objects – human, non-human animal, rigid object etc. The advantage of this method is that – (a) we do not need exact figure; (b) it is scale invariant – thus the camera image can be taken from close or from far; (c) It works even with somewhat deformed shape (for example, if the person is bending slightly), or people with different clothes

The broad steps of our detection algorithm are as follows –
1. Store the features of the object skeleton currently being processed in appropriate cells or arrays.
2. Differentiate between fork points, end points and branch points of the skeletons and keep track of these points' relative positions.
3. Find out the topmost and bottommost end points (T and B, respectively) along the height of the object.

**Detection of the posture–**
Once the skeleton was found, we first tried to determine if the overall shape was "close" to that of a human. For a human being, one can expect the height to width ratio to stay within a certain limit, which will be different for most other animals. To find the height of the skeleton, we found the topmost endpoint and the bottommost end point, and computed the vertical shift. Similarly, to compute the width, we found the left most and rightmost end points and computed the horizontal shift. The ratio of these two shifts defined the height/width ratio.

4. Find the leftmost (L) and rightmost (R) end points of the skeleton.
5. Find the vertical shift, V between T and B, and horizontal shift, H between L and R.
6. Find out the ratio V/H. If the ratio is within a pre-defined range, this is a candidate for detection as human: we set **possibility = 1.** The range is calculated considering data for average human beings. For our purpose, it was taken to be greater than 2.3. That is, if the ratio is greater than 2.3, we set **possibility = 1.** Otherwise, we set **possibility = 0.**

**Detection of the shape** –
Human body is divided into 3 main parts - (a) Head and neck; (b) Trunk; (c) Waist and legs. Thus, the skeleton of a human should have two major fork points: (a) at the neck; and (b) at the waist. Average human beings' neck and waist positions tend to show a relatively invariant ratio with respect to the height of the human being. We used these measurements to identify whether the shape of the object being processed is that of a human being or not. Recall that we detected the topmost and bottommost endpoints of the skeleton in step 3. These intuitively gave the positions of the head and foot. Therefore, a shortest path on the skeleton graph from the topmost point to the bottommost point should ideally traverse the human shape via the neck, trunk and waist. Therefore, the fork points on this shortest path will give an idea about the neck/height and waist/height ratio.

7. Identify the forks between T and B on the shortest path.
8. For each of the fork points do the following
9. Calculate the pixel distance from T to the fork (**shape1**) and from the fork to B (**shape2**).
10. Calculate **shape = shape2 / shape1**.
    This ratio is used to find out the relative positions of the neck and the limbs which help to determine if it is a human or otherwise.
11. If the value of shape is between 5 and 8 identify it as the possible neck of the object; set **shapeneck = 1, otherwise 0**. The figures 5 and 8 have been obtained from average human data collected statistically.
    If the value of shape is between 1 and 2 identify it as the possible waist of the object and set **shapewaist = 1, otherwise 0**. figures 1 and 2 have been obtained from average human data collected statistically.
12. If either shapeneck = 1 or shapewaist = 1 give the object a score of 0.4 and call it **shape_pos**; if both are 1 then give it a score of 0.8.

Detection of Human Being –
13. Sum up the values of possibility and shape possibility. Depending on the sum the following detections were made –
    **Final_Score = possibility + shape_pos**
14. (a) If Final_Score = 0; NO CHANGE
    (b) If Final_Score = 0.4; CHANGE BUT NOT HUMAN BEING
    (c) If Final_Score = 0.8; Generate ALERT, but probably NOT HUMAN BEING
    (d) If Final_Score = 1; Generate ALERT, MOST PROBABLY HUMAN
    (e) If Final_Score = 1.4; Generate ALERT, HUMAN BEING
    (f) If Final_Score = 1.8; Generate ALERT, DEFINITE HUMAN BEING

## 5. Experiments

We have tested our algorithm on several data sets taken under varied conditions using MATLAB-R2009a. Here, we give the results on one data set: detection of a student's appearance in an empty classroom. We have first given the static picture, followed by the dynamic frames; and finally the difference pictures along with the corresponding skeleton of the dynamic object. In the following example (Figure 6), we have first taken some static pictures of the classroom and then using video feed collected some dynamic frames with a student in that same place. The student is moving forward continuously. We have shown 4 such dynamic frames and gathered the difference pictures by comparing between the static picture and the real time dynamic frames. The difference pictures are binary images. The corresponding skeletons of the dynamic objects are then obtained. The final score came out to be 1.4, which is strong enough. We have tested our algorithm on 30 different situations involving men & women wearing different clothes; and also on dynamic frames containing animals, such as dogs, cats; and rigid objects like cars and boxes. We have conducted the experiments under different illumination levels and at different places. We have observed promising results using this approach.

## 6. Conclusion

In this paper, we have described an algorithm for human detection using the measurements of the corresponding skeleton. This algorithm is capable of detecting human beings in different environments, under different lighting conditions. It is deformity tolerant to a significant extent in the sense that bending or twisting does not disable its posture. Many of the existing generative-based object detection/recognition algorithms are seen to be have either very limited modelling power, or they are too complicated to learn, and also tends to be computationally expensive. Our algorithm, however, is simple, effective and efficient. The results are encouraging. There are issues open to further investigation. Very noisy environments (for example, a background that itself changes substantially) generate erroneous difference images. Sudden sharp change in lighting conditions also introduces artifacts. The ratios which we used were obtained after analyzing average human data; so there are possibilities that some humans will fall out of this range. Also, we could not test our algorithm on primates which are close to humans in shape; so we do not know how it will perform. More experiments on these issues can further enhance the efficiency of the algorithm.

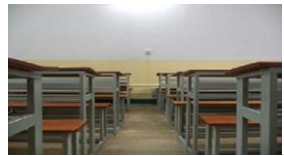

**STATIC PICTURE    PLACE: ST XAVIER'S COLLEGE CLASSROOM**

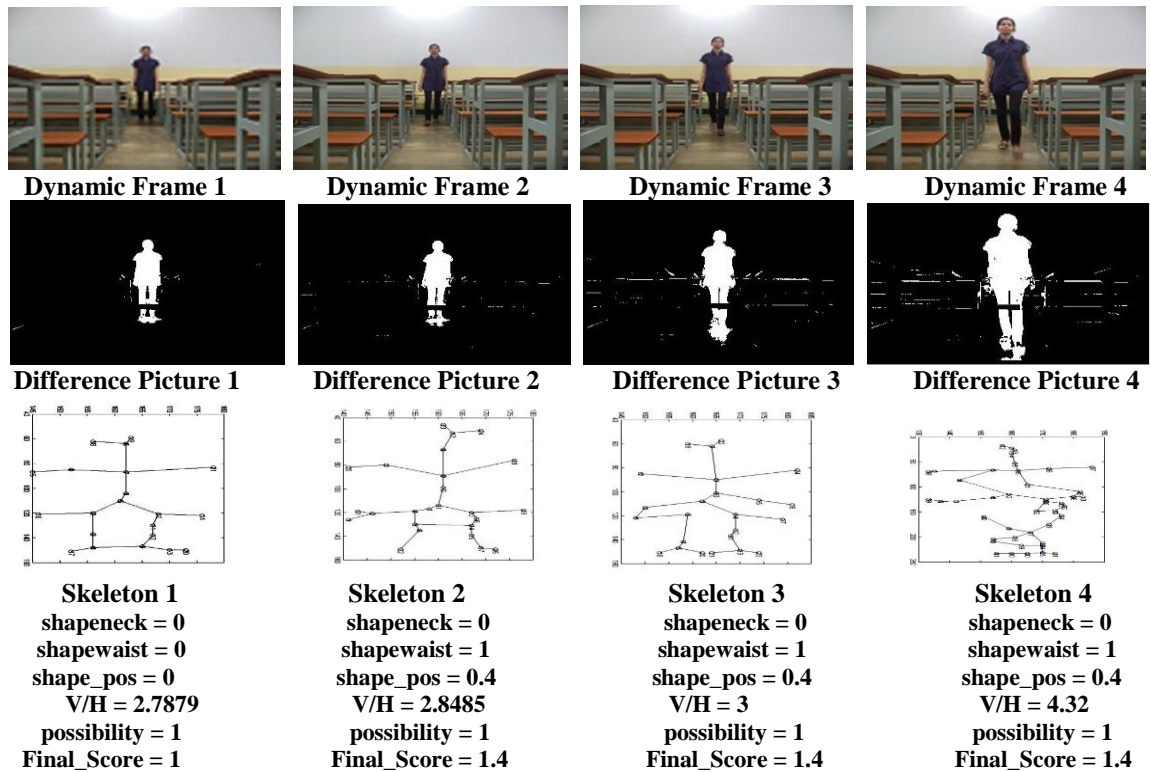

| Dynamic Frame 1 | Dynamic Frame 2 | Dynamic Frame 3 | Dynamic Frame 4 |

| Difference Picture 1 | Difference Picture 2 | Difference Picture 3 | Difference Picture 4 |

**Skeleton 1**
shapeneck = 0
shapewaist = 0
shape_pos = 0
V/H = 2.7879
possibility = 1
Final_Score = 1

**Skeleton 2**
shapeneck = 0
shapewaist = 1
shape_pos = 0.4
V/H = 2.8485
possibility = 1
Final_Score = 1.4

**Skeleton 3**
shapeneck = 0
shapewaist = 1
shape_pos = 0.4
V/H = 3
possibility = 1
Final_Score = 1.4

**Skeleton 4**
shapeneck = 0
shapewaist = 1
shape_pos = 0.4
V/H = 4.32
possibility = 1
Final_Score = 1.4

**Figure 6. Static Image, Dynamic frames, skeletons and scores obtained on one set of data.**

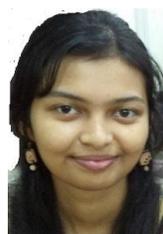
**Dhriti Sengupta** received her B.Sc.(Hons) degree with 1st class in Computer Science from the University of Calcutta, Kolkata, India, in 2012. She is currently pursuing her Master of Science degree in Computer Science from St. Xavier's College (Autonomous), under the University of Calcutta, Kolkata, India. Her current research interests include image processing, network security, and algorithms.

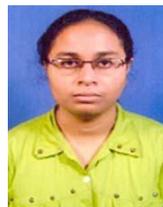
**Merina Kundu** received her B.Sc.(Hons) degree with 1st class in Computer Science from the University of Calcutta, Kolkata, India, in 2012. She is currently pursuing her Master of Science degree in Computer Science from St. Xavier's College (Autonomous), under the University of Calcutta, Kolkata, India. Her current research interests include network security and image processing.

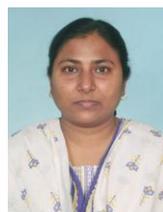
**Jayati Ghosh Dastidar** completed her Master in Engineering (Information Technology) from the West Bengal University of Technology, Kolkata, India, in 2005. She is currently an Assistant Professor in the Department of Computer Science, St. Xavier's College (Autonomous), under the University of Calcutta, Kolkata, India. Her current research interests lie in the field of image processing.